\documentclass[10pt,twocolumn,letterpaper]{article}

\usepackage{iccv}
\usepackage{times}
\usepackage{epsfig}
\usepackage{graphicx}
\usepackage{amsmath}
\usepackage{amssymb}
\usepackage{multirow}

\usepackage[breaklinks=true,bookmarks=false]{hyperref}

\iccvfinalcopy

\def\iccvPaperID{7243}

\ificcvfinal\fi

\begin{document}

\title{Latent-OFER: Detect, Mask, and Reconstruct with Latent Vectors for Occluded Facial Expression Recognition}

\author{Isack Lee, Eungi Lee, Seok Bong Yoo\thanks{Corresponding author} \\
Department of Artificial Intelligence Convergence, Chonnam National University, Gwangju, Korea\\
{\tt\small \{sackda24, 181061, sbyoo\}@jnu.ac.kr}
}

\maketitle
\ificcvfinal\thispagestyle{empty}\fi

\begin{abstract}
Most research on facial expression recognition (FER) is conducted in highly controlled environments, but its performance is often unacceptable when applied to real-world situations. This is because when unexpected objects occlude the face, the FER network faces difficulties extracting facial features and accurately predicting facial expressions. Therefore, occluded FER (OFER) is a challenging problem. Previous studies on occlusion-aware FER have typically required fully annotated facial images for training. However, collecting facial images with various occlusions and expression annotations is time-consuming and expensive. Latent-OFER, the proposed method, can detect occlusions, restore occluded parts of the face as if they were unoccluded, and recognize them, improving FER accuracy. This approach involves three steps: First, the vision transformer (ViT)-based occlusion patch detector masks the occluded position by training only latent vectors from the unoccluded patches using the support vector data description algorithm. Second, the hybrid reconstruction network generates the masking position as a complete image using the ViT and convolutional neural network (CNN). Last, the expression-relevant latent vector extractor retrieves and uses expression-related information from all latent vectors by applying a CNN-based class activation map. This mechanism has a significant advantage in preventing performance degradation from occlusion by unseen objects. The experimental results on several databases demonstrate the superiority of the proposed method over state-of-the-art methods. This code is available at https://github.com/leeisack/Latent-OFER.
\end{abstract}

\section{Introduction}

\begin{figure}
\begin{center}
\includegraphics[width=0.95\linewidth]{"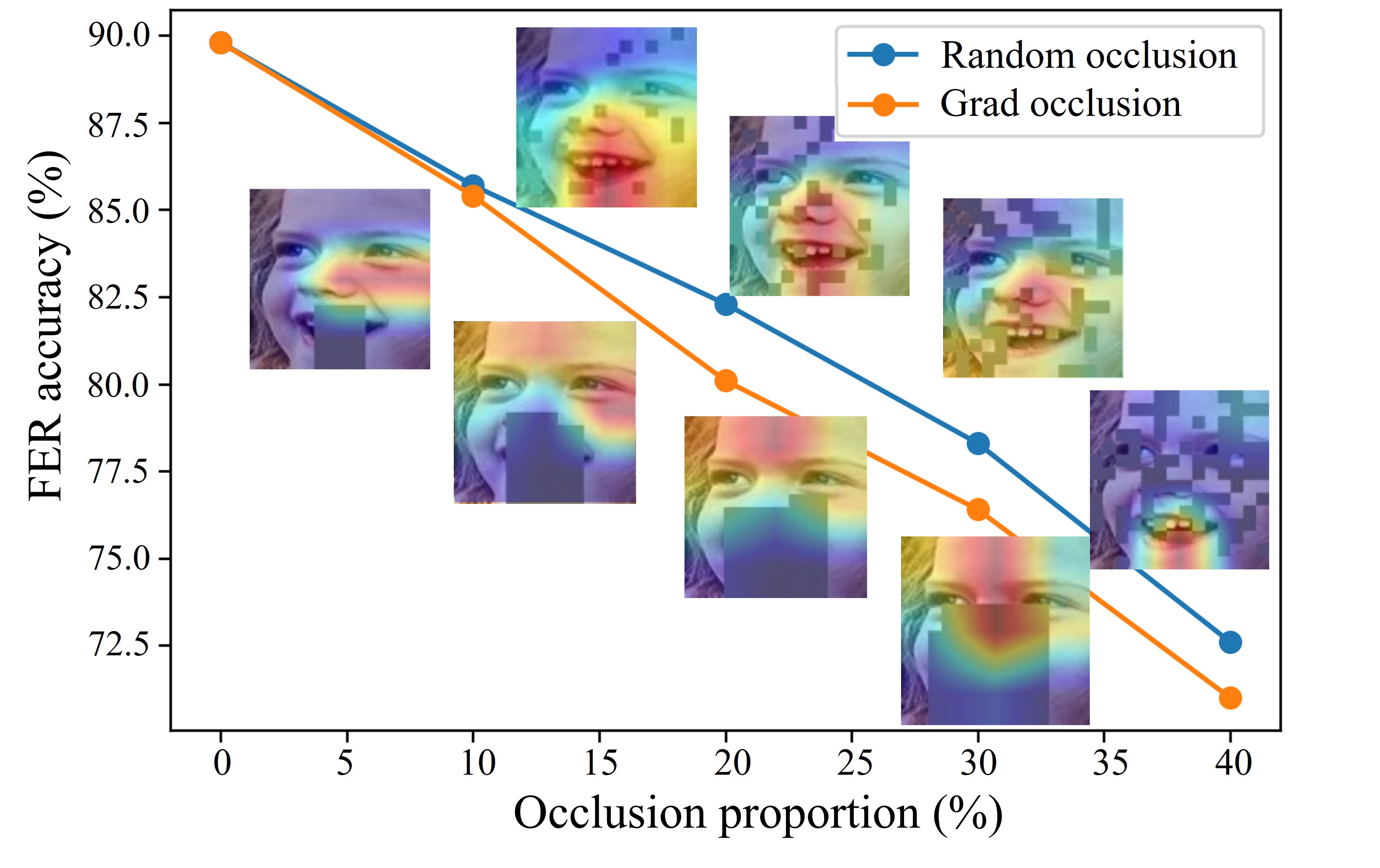"}
\end{center}
\vspace{-14pt}
\caption{Facial expression recognition (FER) performance on RAF-DB according to the occluded proportions.}
\vspace{-20pt}
\end{figure}

Facial expression recognition (FER) has undergone remarkable advancements in recent years and is now widely used across various industries. However, the ability of FER models in the presence of occlusion remains a challenge. Figure 1 illustrates the accuracy of the previous state-of-the-art model for FER \cite{DAN} for various occlusion proportions. In this study, we evaluated the robustness of the typical model in recognizing facial expressions using two types of occlusions: random sampling occlusion and grad occlusion. Random sampling occlusion divides the entire image into 196 patches and randomly masks them according to the proportion. Grad occlusion processes an image with an intentionally occluded area that affects FER using gradient-weighted class activation mapping (Grad-CAM) \cite{Grad_CAM}. This study reveals a more substantial decrease in performance with the second type of occlusion, particularly when the occluded area is crucial for accurate FER. This finding has been objectively measured using Grad-CAM \cite{Grad_CAM}.

Previous studies on FER \cite{17,5,8,10,9,6,25,DACL,3,24,12,7,26,20,23,15,16,14,4,19,11,13,RUL,EAC,18} have not given adequate attention to the influence of occlusions. However, addressing this challenge is crucial for enhancing the practical applications of FER in real-world scenarios. Although research on occluded FER (OFER) is relatively scarce, its importance is increasingly recognized \cite{survey_occ}. 

Currently, several approaches address OFER. The occlusion-robust feature extraction approach \cite{OADN,Wangs} aims to identify an occlusion-insensitive and discriminative representation, but it is challenging because the types and locations of occlusion are often unknown. The sub region analysis approach \cite{attention_cnn,RAN} divides regions based on facial landmarks and uses attention mechanisms to focus on crucial areas. However, the inability to detect facial landmarks due to occlusions can lead to errors in the FER process. The unoccluded image network assist approach \cite{Pans,Xias} uses two distinct networks: one trained on unoccluded images and the other trained on occluded images. This approach leverages unoccluded images as privileged information to assist in expression recognition in the presence of occlusion. It is unsuitable in real-world situations because it cannot differentiate between occluded and unoccluded images.

Hence, the proposed approach is the occlusion recovery-based approach, which aims to transform occluded images into complete images through a deocclusion process. We propose the deocclusive autoencoder for reconstructing facial images. The deocclusive autoencoder can be functionally divided into an occlusion detector and reconstruction module. The occlusion detector uses the vision transformer (ViT) support vector data description (SVDD) for the reconstruction module. This approach enables the model to detect occlusion caused by unseen objects, an essential step in accurately generating deoccluded facial images. The reconstruction network consists of the ViT structure and convolutional neural network (CNN) structure for facial image reconstruction. 
We leverage the strengths of the ViT in generating realistic facial images despite varying poses and further refine them using the CNN. We call it the hybrid reconstruction network. The hybrid reconstruction network enhances FER performance by generating deoccluded images that express detailed and vivid facial expression attributes. This enhancement is achieved by incorporating a self-assembly layer and semantic consistency loss.
In contrast, previous work on image reconstruction has primarily focused on achieving naturalness, which can result in dull facial expressions. Additionally, we use informative ViT-latent vectors obtained from the reconstruction process. We combine the CNN features and ViT-latent vectors for enhanced facial expression predictions. The main contributions in this work are summarized as follows:

• We propose an expression-relevant feature extractor that uses spatial attention to assign a higher weight to specific facial features, allowing us to identify critical positions for FER. We can retrieve expression-relevant latent vectors from the ViT-latent space to extract valuable information using these positions as critical values.\\
•We propose ViT-SVDD, a patch-based occlusion detection module optimized for ViT-based networks. As a self-supervised local classifier, the ViT-SVDD module is trained only on latent vectors of unoccluded facial images. This method accurately classifies occlusions caused by unseen objects for subsequent reconstruction.\\
• We propose a hybrid reconstruction network that combines the strengths of the ViT and CNN architectures with a self-assembly layer and semantic consistency loss to generate facial images naturalness and rich in expression. This approach enhances the quality of deoccluded images and improves the accuracy of FER in challenging conditions.

\section{Related Work}
\subsection{Occluded Object Detection}

Several early studies \cite{37} have attempted to detect occlusions in images for OFER tasks. The conventional methods provide location information about the occlusions and are called occlusion-aware methods \cite{attention_cnn,RAN}. However, these methods are not practical for real-world scenarios. Another approach \cite{WGAN} is to train models using synthesized images with various occlusions to detect the occluded positions. Although this approach can determine occlusion positions, it requires a diverse set of object images, which can be challenging to obtain. Moreover, the performance of these models \cite{attention_cnn,WGAN,RAN} degrades when they encounter unseen objects as input. We drew inspiration from anomaly detection tasks \cite{39,40,41,38} with one-class classification algorithms \cite{45,46,43,44,42} and proposed a model that learns only from unoccluded datasets and can generalize well to unseen datasets.

\begin{figure*} 
\begin{center}
\includegraphics[width=0.95\linewidth]{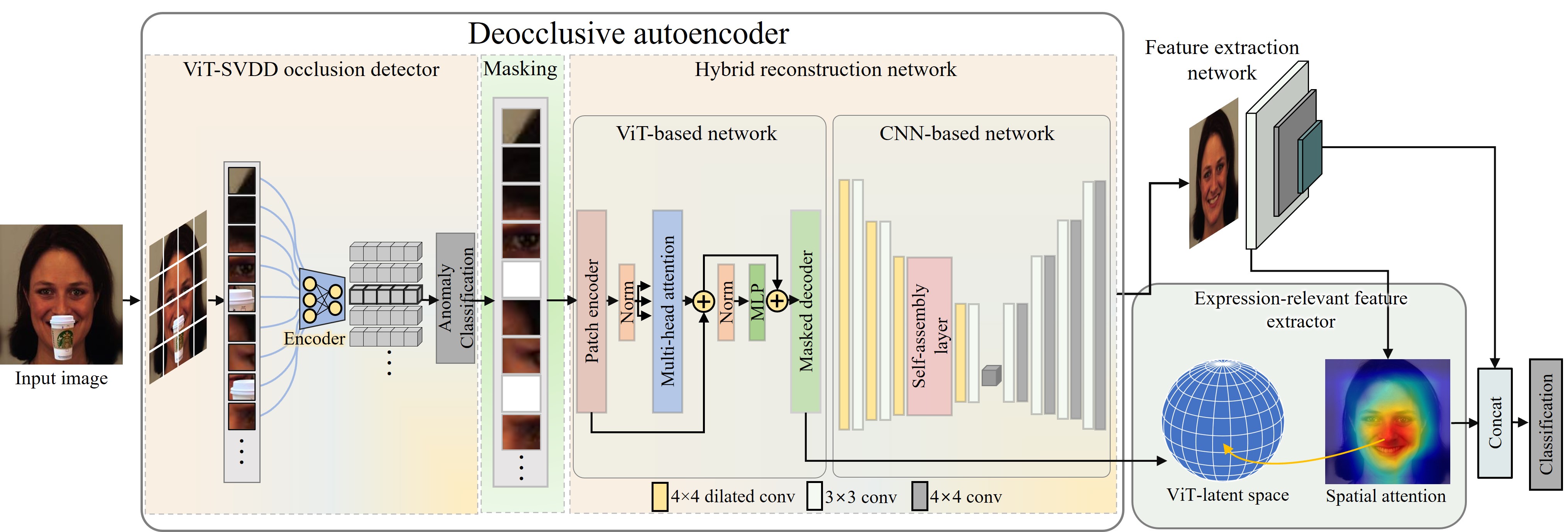}
\end{center}
\vspace{-10pt}
\caption{Framework of the Latent-OFER that creates a deoccluded image that erases the occlusion. In this process, the expression-relevant latent vectors are extracted by mapping the vision transformer (ViT)-latent vector and the convolutional neural network (CNN) class activation map. Latent-OFER predicts facial expressions that combine CNN-based features with specific VIT-based latent vectors.}
\vspace{-10pt}
\label{fig:short}
\end{figure*}

\subsection{Occluded Facial Expression Recognition}
We have no prior knowledge of where occlusions may appear on a facial image or how large or complex they might be. These occlusions can significantly reduce the accuracy of FER by either increasing the intra-class variability or inter-class similarity. Four categories address this issue: occlusion-robust feature extraction, subregion analysis, unoccluded image network-assisted, and occlusion recovery.

\textbf{Occlusion-robust feature extraction approach.} This approach aims to extract features less affected by occlusions while maintaining discriminative capability. Wang et al. \cite{Wangs} used self-supervised and contrastive learning to explore robust representations with synthesized occlusions.

\textbf{Subregion analysis approach.} This approach excludes the occluded parts from recognition. By focusing only on the unoccluded facial parts, the influence of occlusion on recognition performance can be reduced. Li et al. \cite{attention_cnn} proposed a gate unit with an attention mechanism that allows the model to focus on informative unoccluded facial areas.

\textbf{Unoccluded image network-assisted approach.} This approach employs unoccluded facial images as guidance to assist with OFER. Pan et al. \cite{Pans} trained two deep neural networks from occluded and unoccluded facial images. Then, the unoccluded network guides the occluded network. Xia et al. \cite{Xias} measured the complexity of unoccluded data using distribution density in a feature space. The classifier can be guided by unoccluded data and subsequently leverage more meaningful and discriminative samples.

\textbf{Occlusion recovery-based approach.} This approach uses face completion as a basis to deal with occlusion. To reconstruct occluded faces, Lu et al. \cite{WGAN} suggested using a WGAN consisting of an autoencoder-based generator and discriminators. By reconstructing the occlusion, these approaches provide necessary information while avoiding the interference of noisy information and obtaining informative appearance. Therefore, we adopt an image-inpainting technique as a recovery-based approach, to deal with occlusion. A notable point of differentiation between our method and existing approaches lies in our specific emphasis on the reconstruction of occluded regions while simultaneously retaining the facial expression information. Furthermore, we employ ViT-based latent vectors extracted throughout the reconstruction process to enhance the performance of FER.

\subsection{Image inpainting}
Several approaches based on CNN have been developed to generate semantically coherent content. Pathak et al. \cite{pathak2016context} employed context encoders to generate plausible features for the area requiring restoration. Yu et al. \cite{yu2018generative} introduced a contextual attention module to refine by referencing the surrounding features. Song et al. \cite{song2018contextual} used a patch-swap layer that replaces each patch within the masking areas of a feature map with the most comparable patch in the unoccluded areas. Liu et al. \cite{liu2019coherent} developed a coherent semantic attention layer to ensure semantic relevance between the swapped features.

In recent years, transformer-based methods have also significantly contributed to computer vision. The ViT \cite{ViT} achieves better results for image recognition than CNN-based methods. Furthermore, MAE \cite{MAE} demonstrated that transformer-based models are well-suited for image-inpainting tasks. However, transformer-based algorithms have a limitation in that they may produce blurry results when restoring large block areas \cite{MAE}. Another work \cite{MAT} presents MAT, a transformer-based model for large-hole inpainting. Nevertheless, in cases where the eyes or mouth are fully occluded, the ViT cannot generate images in detail because it is significantly less biased toward local textures \cite{propertiesViT}. Therefore, the challenge remains.

\section{Proposed method}
As shown in Figure 2, we propose multi-stage approach to address OFER, involving detecting, masking, and reconstructing occlusions to recognize facial expressions. The proposed approach enhances recognition accuracy through cooperative learning ViT-latent vectors extracted from the image reconstruction process and the existing CNN features. We divided the facial image into patches, classified each patch as occluded or unoccluded, and reconstructed the occluded patches to be deoccluded. Subsequently, we leveraged the reconstructed image and expression-relevant latent vectors to predict facial expressions.
\subsection{Occlusion detection module: ViT-SVDD}
General object detection and segmentation models may not be suitable for real-world scenarios because they may be unable to detect unseen objects. To address this limitation, we use one-class classification, which is often employed in anomaly detection. This approach makes it possible to classify unused occlusion during training through self-supervised learning, which only uses unoccluded patches for learning, providing a more effective solution for real-world applications.

One-class classification methods for classifying normal or abnormal can operate at various levels of granularity, ranging from low-level anomaly detection \cite{41} at the pixel level to high-level anomaly detection \cite{44} at the image level. Detection and classification are performed based on the size of the unit, which can be customized to suit the user’s needs. We used a ViT-based reconstruction method; thus, we propose a middle-level anomaly detector specifically optimized for ViT. We divided the image to match the size of the ViT patch and created ViT-latent vectors. These patches are encoded with informative features to produce ViT-latent vectors. To generate the smallest feature space for unoccluded patches, we used the deep SVDD algorithm \cite{43}. One-class deep SVDD employs quadratic loss to penalize the distance of every network representation. This objective is define as\\
\vspace{-10pt}
\begin{equation}
    min\frac{1}{n}\sum_{i=1}^{n}\left\| \Phi\left ( x_{i}; \mathcal{W} \right ) - c\right\|^{2} + \frac{\lambda }{2}\sum_{\textit{l}=1}^{L}\left\| \textsl{w}^{\textit{l}}\right\|_{F}^{2}.
\end{equation}
\vspace{-10pt}

\noindent where $n$ denotes number of training data, $L$ denote number of layer, $\mathcal{W}$ denotes set of weights $\mathcal{W}$ $=\left\{ \textsl{w}^{1},...,\textsl{w}^{L}\right\}$ and $c$ represents a hypersphere characterized by the center. The first term induces the features of all normal images to converge at the center point $c$, whereas the last term is a weight decay regularizer on the network parameter $W$ with hyperparameter $\lambda\ >0$, where $\left\|\cdot \right\|_{F}$ denotes the Frobenius norm.  Eq (1) simply employs a quadratic loss for penalizing the distance of every network representation $\Phi\left ( x_{i}; \mathcal{W} \right )$ to $c\in\mathcal{F}$, Where $\mathcal{F}$ is output feature space. The network learn parameters $\mathcal{W}$ such that data points are closely mapped to $c$ of the hypersphere. To determine whether a patch is occluded, we calculated the distance between the new input information and the center $c$ of the feature space for each patch. If the distance exceed the pre-defined radius, the corresponding patch is classified as occluded and masked. The optimal value of the radius is automatically determined in the SVDD procedure. Through this process, occlusion patch detection is possible for unseen objects. The proposed ViT-SVDD approach allows for validating the performance of synthetic images with occlusion patch annotations. By detecting occlusion patches, we can improve the accuracy of the reconstruction method and make it more suitable for real-world applications.

\subsection{Image reconstruction module: hybrid reconstruction network}
The facial image reconstruction process employs an occlusion-masked image generated by the occlusion detector. The hybrid reconstruction network is designed to cooperate by fusing ViT-based and CNN-based networks. Through this mechanism, we leverage both the strengths of ViT and CNN. The ViT-based method employs $16\times16$ patches as input image; however, we used the outputs of the occlusion detector as input because the image has already been partitioned into patch units by the occlusion detector.
\vspace{-10pt}
\subsubsection{Network structure}
\quad The ViT-based approach encodes the input patches and positively embeds all tokens. The occluded patch reconstruction is achieved through correlation with other patches. The ViT has low inductive bias and a high degree of freedom \cite{propertiesViT}, enabling it to generate credible images despite diverse occlusion shapes, positions, and facial poses. As discussed in Section 2.3, the ViT-based approach may sometimes not supply detailed results. To address this limitation, we combined the ViT and CNN. The network consists of a U-Net architecture. Additionally, we added a self-assembly layer inside the encoder to generate a detailed representation. This multi structure approach effectively combines the strengths of the ViT and CNN-based networks to generate high-quality facial image reconstructions that represent facial expressions well.
\vspace{-10pt}

\begin{figure}
\vspace{-15pt}
\begin{center}
\includegraphics[width=0.8\linewidth]{"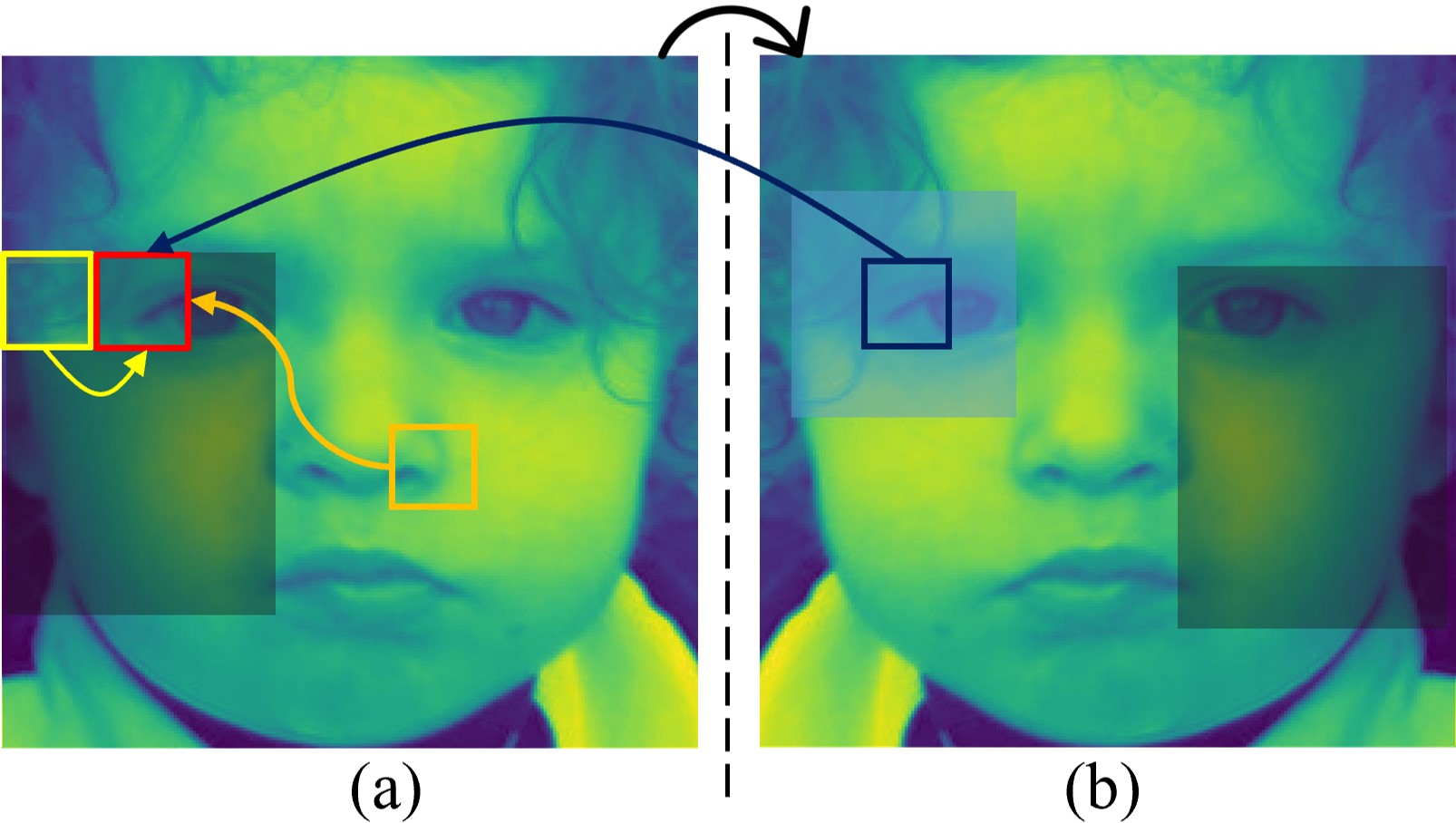"}
\end{center}
\vspace{-10pt}
\caption{Illustration of the self-assembly operation. (a) Feature map in the self-assembly layer. (b) Flip of (a). The patch in masked region is generated by combining three patch information.}
\label{fig:long}
\label{fig:onecol}
\vspace{-10pt}
\end{figure}

\subsubsection{Self-assembly layer}
\quad We implemented a self-assembly layer to improve image reconstruction for FER. We reconstructed masked regions like \cite{MAE} but made enhancements specific to facial images. Based on the concept that the left and right characteristics of a person’s face are symmetrical\cite{sym3,sym1,sym2}, we used the feature information present in the corresponding location of the horizontally flipped image when reconstructing a masked region. We expanded the range of candidate patches used in the generation process by incorporating information from three sources, the previously generated patch, the most similar patch in the unmasked region, and patches located in the corresponding position of the horizontally flipped image. In this process, the masked region contains the reconstruction results with the ViT network. We assigned weights to each based on the similarity value with the current patch. The weight calculation is based on a cross-correlation metric:
\begin{equation}
    \mathit{S(\mathit{p},\mathit{p}_x)}=\frac{\left<\mathit{p},\mathit{p}_x \right>}{\left\|\mathit{p}\right\|\cdot\left\| \mathit{p}_{x}\right\|},
\end{equation}
where $\mathit{p}$ represents a patch in the masked region, and the patch $\mathit{p}_x$ is the comparison target. The similarity value between $\mathit{p}$ and $\mathit{p}_x$ is denoted by $\mathit{S}$.

The self-assembly operation is defined in Eq (3), which generates a patch value $\mathit{p}_i$. Figure 3 depicts the operation process, where $\mathit{p}_s$ denotes the patch value symmetrically positioned with respect to $\mathit{p}$ and averaged by considering the peripheral patches. The $\mathit{S}_{sym}$ is calculated as Eq (2) through $\mathit{p}_i$ and $\mathit{p}_{s_i}$. In addition, $\mathit{p}_k$ is the most similar patch to $\mathit{p}$ among the unmasked region, and $\mathit{S}_{known}$ is calculated as $\mathit{S}(\mathit{p}_{i},\mathit{p}_{k_i})$. Further, $\mathit{p}_{i-1}$ represents the previously generated patch, and $\mathit{S}_{i-1}$ is obtained by $\mathit{S}(\mathit{p}_{i},\mathit{p}_{i-1})$. The similarity value $\mathit{S}$ is normalized to be used as the weight.

In Figure 3, the red patch in (a) is $\mathit{p}_i$, which is a combined result of $\mathit{p}_k$ (the orange patch), $\mathit{p}_{i-1}$ (the yellow patch), and $\mathit{p}_s$ (the blue patch in (b)). Because $\mathit{p}_1$ has no prior generated patch, $\mathit{S}_0$ is zero. In certain cases, such as side-face images, the patches in symmetrical positions may not be relevant for generating the patch. Thus, the value of $\mathit{S}_{sym}$ in this situation is minimal and is rarely used to generate the patch.
\begin{align} \label{eqn:my_equation}
  \begin{split}
	\mathit{p}_i=\frac{S_{sym}}{S_{sym}+S_{known}+S_{i-1}}\times \mathit{p}_{s_{i}}\\
	{+} \frac{S_{known}}{S_{sym}+S_{known}+S_{i-1}}\times \mathit{p}_{k_{i}}.\\
        {+} \frac{S_{i-1}}{S_{sym}+S_{known}+S_{i-1}}\times \mathit{p}_{i-1}\\
  \end{split}
\end{align}

\vspace{-7pt}

\subsubsection{Objective}
\quad The aim of image reconstruction is to fill in the masked portion to provide supplementary information for FER. In pursuit of this, we incorporated a semantic consistency loss that enables optimizing the task while maintaining the reconstruction loss $L_{re}$, consistency $L_{c}$ \cite{liu2019coherent}, feature patch discriminator $L_{d_f}$ \cite{56}, and patch discriminator $L_{d}$ \cite{57}.

The semantic consistency loss $L_{sc}$ emphasizes facial expression attributes. The $L_{sc}$ has the effect of reducing intra-class variability and can be defined as:
\vspace{-8pt}

\begin{equation}
    L_{sc}=\sum_{c=1}^{7}p_c\left ( z_{gt} \right )log\left ( p_c\left ( z_{rec} \right ) \right ),
\end{equation}

\noindent where c represents seven basic expressions, $p_c\left ( z_{gt} \right )$ denotes the predicted probability of $c$ in the ground-truth image, and $p_c\left ( z_{rec} \right )$ indicates the predicted probability of $c$ in the reconstruction result. The predicted probability distributions are obtained via the pretrained FER network. During training, the overall loss function is defined as: 
\vspace{-10pt}

\begin{equation}
L = \lambda _{re}L_{re}+\lambda _{c}L_{c}+\lambda _{sc}L_{sc}+\lambda _{d}(L_{d}+L_{d_{f}}),
\end{equation}

\noindent where $\lambda_{re}, \lambda_c, \lambda_{sc}, \lambda_d$ denote the trade-off parameters for the reconstruction, consistency, semantic consistency, discriminator losses, respectively. Furthermore, the FER network is trained with the same feature extraction architecture using probability distributions about the ground-truth label and prediction of FER. 

\subsection{Facial expression recognition network}

The proposed FER network is designed as an attention-based model for predicting facial expressions. We employed spatial and channel-attention mechanisms \cite{cbam}. We obtained the refined feature map and CAM using the attention-based model. In addition, we obtained the expression-relevant latent vectors from the ViT using the CAM. As depicted in Figure 2, the Latent-OFER cooperatively uses a CNN-based feature and ViT-based latent vectors. Thus, the model performs better.

\begin{figure}
\begin{center}
\includegraphics[width=1.0\linewidth]{"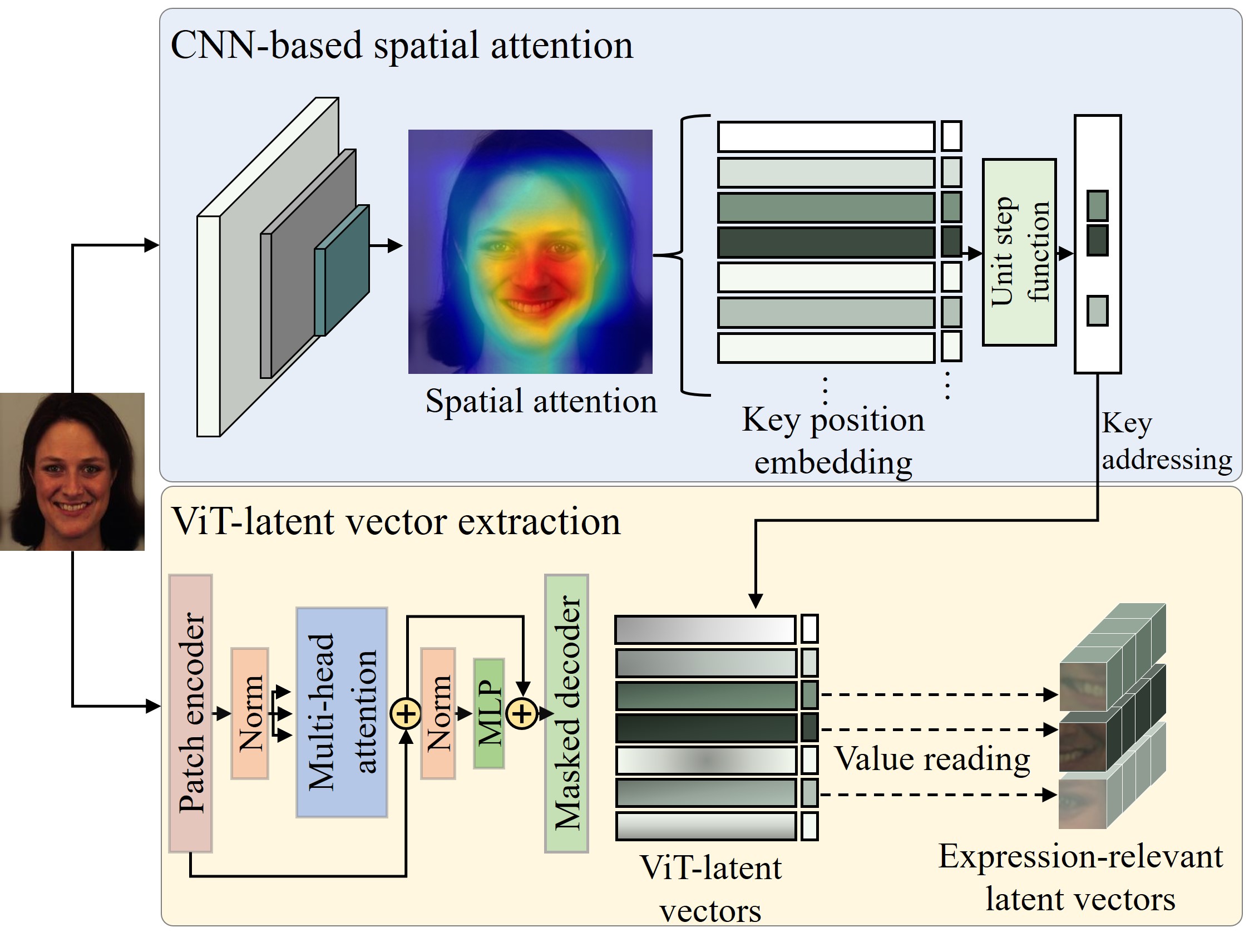"}
\end{center}
\vspace{-10pt}
\caption{Expression-relevant vision transformer (ViT)-latent vector extraction method.}
\vspace{-10pt}
\end{figure}

\textbf{Expression-relevant ViT-latent vectors.}
The proposed method employs only expression-relevant latent vectors rather than the entire latent space to improve FER performance. During the reconstruction process, ViT-based latent vectors are extracted by embedding the input image. We used a CAM to identify spatially significant areas within the image for FER, and the class activation map is generated through a CNN.\\
\vspace{-10pt}

The position of the area is stored as a key, and the attention weight for each space is recorded. The key of the region where the weight of the spatial attention exceeds the top 50\% is used. This key is retrieved from the entire ViT-latent vectors, and the corresponding value is read. The activation map is used to identify expression-relevant latent vectors, as presented in Figure 4. This process enables the selection of positions that are relevant to FER while avoiding extraneous details, such as appearance information irrelevant to expressions, which can increase inter-class dissimilarity and lead to more accurate and effective learning results.\\
\vspace{-10pt}

In cases where patch detection fails, the occluded patch’s latent vectors are used for training and inference. However, spatial attention is not focused on the occluded area, and the latent vector of the occlusion patch is neither searched nor used for training and inference.The proposed extractor is not significantly affected in this situation.

\section{Experiments}
In this part, we describe three FER benchmark datasets and several occluded FER datasets used to evaluate this model and explain the results. We demonstrate that the proposed method performs better on both the FER task and occluded FER task. Last, an ablation study demonstrates how each model element contributes to the final performance.

\subsection{Dataset}
RAF-DB \cite{RAF} is a large-scale facial expression database with around 30,000 images. We used data on seven basic expressions, 12,271 images as the training set, and 3,068 images as the testing set.

AffectNet \cite{AffectNet} is the largest facial expression database with annotations. We used seven basic expressions facial images, about 287,568 training and 3,500 testing images.

KDEF \cite{KDEF} is a set of 4,900 images of facial expressions. Each expression is viewed from five viewpoints. 

Syn-AffectNet and Syn-RAF-DB consist of FER benchmark datasets that synthesize a real object to occlude.

Occlusion-AffectNet and Occlusion-RAF-DB are real-world occlusion datasets selected from the AffectNet validation set and RAF-DB testing set by Wang et al. \cite{Wangs}.

FED-RO dataset is also a real-world occlusion dataset collected by Li et al. \cite{attention_cnn}. It contains 400 images labeled with seven basic emotions. We trained the proposed model using RAF-DB and AffectNet when testing the FED-RO datasets, following the protocol suggested in \cite{attention_cnn}.

\begin{table}
\centering
\resizebox{0.9\columnwidth}{!}{%
\begin{tabular}{|c|c|c|c|}
\hline
                       & Accuracy (\%) & Precision (\%) & Recall (\%) \\ \hline
One class SVM \cite{43} & {\color{blue}91.1}     & {\color{blue}90.2}      & 89.4   \\ \hline
Patch-SVDD \cite{38}    & 85.2     & 80.1      & {\color{blue}94.1} \\ \hline
ViT-SVDD               & {\color{red}98.3}   & {\color{red}94.1}    & {\color{red}98.7}  \\ \hline
\end{tabular}
}\vspace{2pt}
\caption{Comparison of occlusion patch detection performance.}
\vspace{-5pt}
\end{table}

\begin{figure}
\begin{center}
\includegraphics[width=0.9\linewidth]{"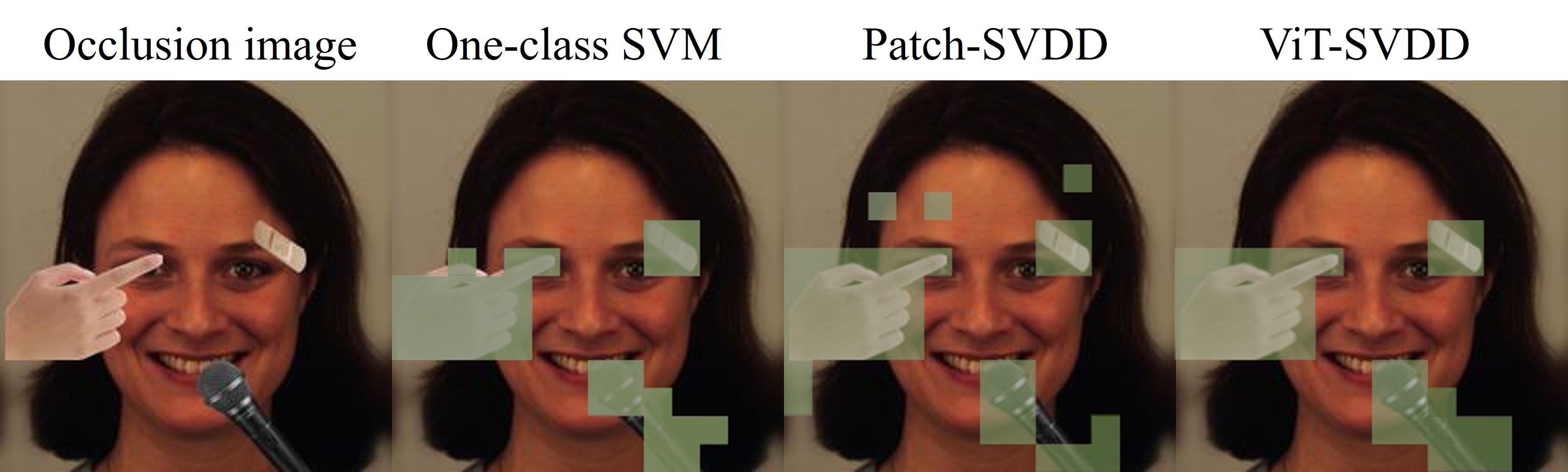"}
\end{center}
\vspace{-10pt}
\caption{Occlusion patch classification results.}
\vspace{-15pt}
\end{figure}

\subsection{Implementation details}
We trained a self-supervised occlusion detection module with the KDEF dataset \cite{KDEF} synthesized by randomly copy-pasting objects such as hands and cups for occlusion. We used a ViT-based network pretrained with ImageNet \cite{imagenet} and a CNN-based network trained on several FER training datasets \cite{RAF,AffectNet} for detailed representation in the image reconstruction module. The FER network uses ResNet-18 \cite{ResNet} backbone architecture. It is pretrained on MS-Celeb-1M \cite{MS-celeb}. The trade-off parameters in hybrid reconstruction module are set as $\lambda_{re}$=1, $\lambda_{c}$=0.01, $\lambda_{sc}$=1, $\lambda_{d}$=0.002. The experimental source code is implemented with PyTorch, and the models are trained with a GTX-3090 GPU.

\subsection{Comparison of occlusion detection module}
 The proposed ViT-SVDD detects occlusion for each patch using a suitable detector for ViT-based image reconstruction. We compared the module with existing one-class classification methods trained solely on unoccluded images to evaluate its effectiveness. We used a test set consisting of occluded and unoccluded images.

Patch-SVDD, which trains in a hierarchical structure from small pixels to large pixels and segments anomaly positions into small pixel units, was also included in the comparison. The proposed module achieved the highest occlusion detection accuracy among all methods, as demonstrated in Table 1.

Based on these results, the proposed method is particularly well-suited for the patch-by-patch processing rather than being the optimal solution for all anomaly detection tasks. While several proposed anomaly detection methods are suitable for various units, such as whole images and pixels, this approach is specifically designed to fit the ViT-SVDD framework and detect anomalies in a patch-specific manner. As illustrated in Figure 5, the results demonstrate improved performance in patch-specific occlusion detection, indicating that the proposed method is best suited for special situations that require this processing type. 

\subsection{Comparison of reconstruction module}

\begin{figure}
\begin{center}
\includegraphics[width=0.9\linewidth]{"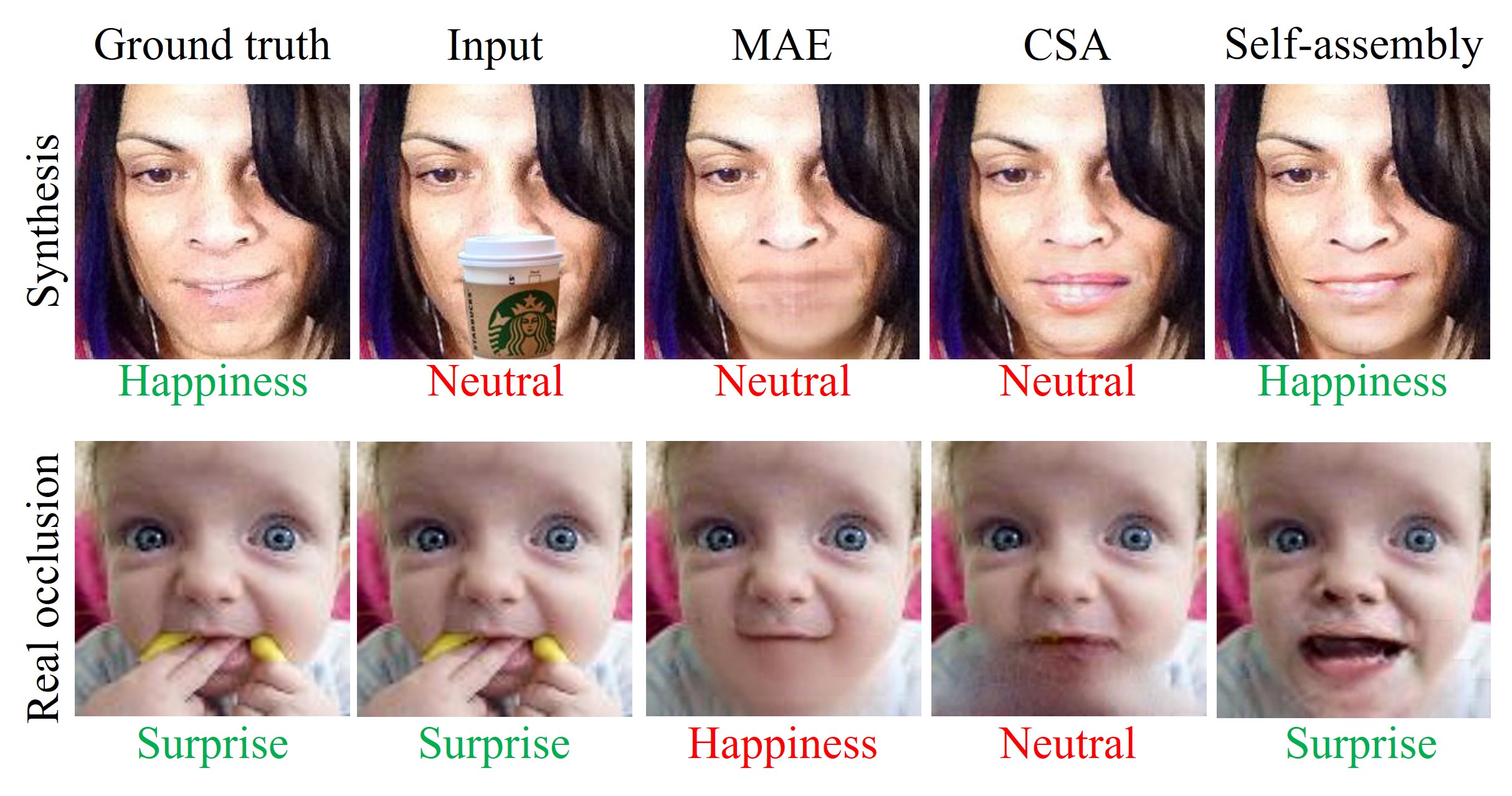"}
\end{center}
\vspace{-13pt}
\caption{Qualitative comparison of synthesis and real occlusion. Facial expression recognition prediction results according to image reconstruction. Labels highlighted in green indicate matching the correct expression, whereas red indicates a misprediction.}
\vspace{-5pt}
\end{figure}

The hybrid reconstruction network is compared to MAE \cite{MAE} and CSA \cite{liu2019coherent}. All reconstruction results are the direct output of the model without post processing. Figure 6 compares the results of deocclusion for synthesis and real occlusion on RAF-DB and AffectNet. As displayed in Figure 6 and Table 2, we present qualitative and quantitative comparisons and the FER result. The self-assembly layer enlarge the facial expression information and increase the number of patch candidates involved in the generation process to reconstruct visually pleasing and natural, and semantic consistency loss induces rich in expression of facial images, providing an advantage for FER. While other comparative models generate reasonable content, the proposed approach offers a superior representation of FER.

\begin{table}
\centering
\resizebox{0.89\columnwidth}{!}{%
\begin{tabular}{|c|c|c|c|}
\hline
              & MAE \cite{MAE} & CSA \cite{liu2019coherent} & Self-assembly \\ \hline
Accuracy (\%) & 72.6         & 75.6         & 77.3         \\ \hline
\end{tabular}
}
\vspace{5pt}
\caption{Facial expression recognition prediction results according to image reconstruction on Syn-RAF-DB.}
\vspace{-3pt}
\end{table}
As presented in Figure 7, the first row shows a partially masked facial image, ground truth image, the corresponding reconstruction results of MAE, CSA and self-assembly. The second row is depicted for better visibility of the influence on FER. The proposed method display the highest probability of a ground truth label.

\begin{figure}
\vspace{-10pt}
\begin{center}
\includegraphics[width=0.95\linewidth]{"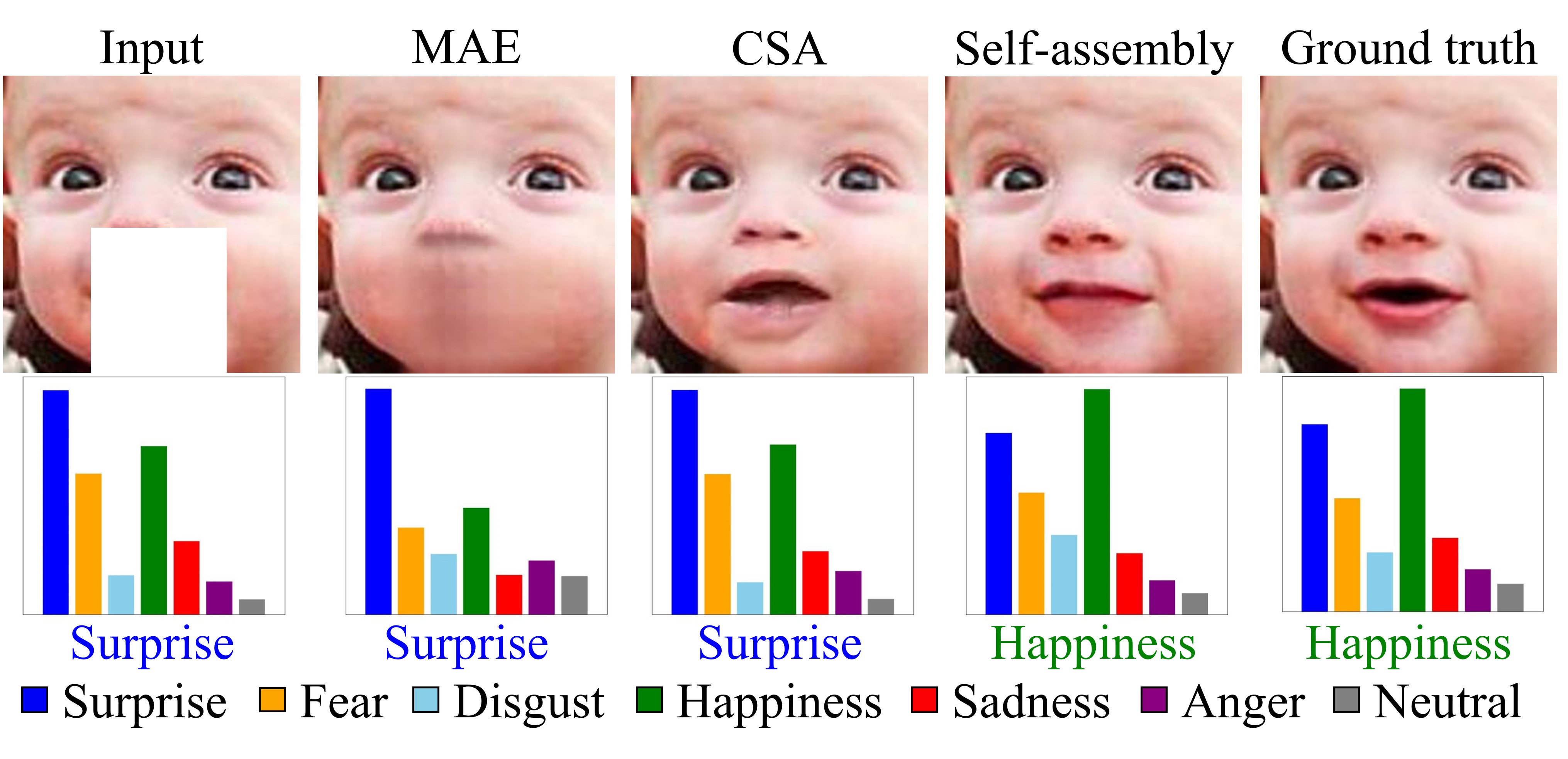"}
\end{center}
\vspace{-15pt}
\caption{Comparison of reconstruction results and the results of each expression probability.}
\end{figure}

\subsection{Comparison of FER accuracy in occlusion}
\textbf{Comparison with Typical FER-model.} As listed in Table 3, we compared the proposed model with several state-of-the-art methods for FER on the AffectNet (C7), RAF-DB, and KDEF datasets, all available as open source. We conducted experiments by separately testing these models with synthesized occlusion images and the original dataset for each dataset. In the field of FER, recent models \cite{DACL,DAN,RUL,EAC, ryumina2022search, kollias2021distribution} demonstrated satisfactory performance on the original dataset. However, they fail to achieve acceptable accuracy on occluded datasets. In contrast, the proposed model was designed to be robust to occlusion and outperformed other models in terms of accuracy on occluded images. Moreover, the proposed model also achieved nearly the best performance on the original images compared to typical models.

\begin{table}
\centering
\resizebox{0.85\columnwidth}{!}{%
\begin{tabular}{|c|c|c|c|c|c|c|}
\hline
Method        & \begin{tabular}[c]{@{}c@{}}Affect\\ Net\end{tabular}                   & \begin{tabular}[c]{@{}c@{}}Syn-\\Affect\\ Net\end{tabular}               & \begin{tabular}[c]{@{}c@{}}RAF\\ DB\end{tabular}                      & \begin{tabular}[c]{@{}c@{}}Syn-\\ RAF \\DB\end{tabular}                  & KDEF                        & \begin{tabular}[c]{@{}c@{}}Syn-\\ KDEF\end{tabular}                    \\ \hline
DACL \cite{DACL} & 62.5                        & 49.6                        & 88.8                        & 76.0                          & {\color{blue} 87.9} & 68.9                        \\ \hline
RUL \cite{RUL}  & 62.6                        & {\color{blue}52.2} & 88.9                        & {\color{blue} 77.1} & 83.1                        & 51.2                        \\ \hline
DAN \cite{DAN}   & {\color{blue} 63.8} & 51.1                        & {\color{red} 89.7} & 76.1                        & 86.8                        & {\color{blue} 70.8} \\ \hline
EAC \cite{EAC}  & 63.2                        & 54.9                        & 88.0                          & 78.6                        & 87.5                        & 67.6                        \\ \hline
\begin{tabular}[c]{@{}c@{}}Latent\\-OFER\end{tabular}    & {\color{red} 63.9} & {\color{red} 56.1} & {\color{blue} 89.6} & {\color{red} 80.1} & {\color{red} 88.3} & {\color{red} 86.7} \\ \hline
\end{tabular}
}
\vspace{3pt}
\caption{Accuracy (\%) comparison to the state-of-the-art results on the various facial expression recognition datasets.}
\vspace{-10pt}

\end{table}

\textbf{Comparison with the OFER-model.} We compared the performance of the proposed FER model targeting occlusion with other state-of-the-art models using FED-RO. As the code for other models is unavailable, we used the reported accuracy from the respective papers. Table 4 reveals that the proposed model achieved an accuracy of 71.8\%$p$ with the default setting, a new state-of-the-art performance for this dataset, to the best of our knowledge.

\begin{table}
\centering
\resizebox{\columnwidth}{!}{
\begin{tabular}{|c|c|c|c|c|c|c|c|}
\hline
       & \begin{tabular}[c]{@{}c@{}} gACNN\\ \cite{attention_cnn}\end{tabular} & \begin{tabular}[c]{@{}c@{}} Pan's \\\cite{Pans}\end{tabular} & \begin{tabular}[c]{@{}c@{}} Xia's \\\cite{Xias}\end{tabular} & \begin{tabular}[c]{@{}c@{}} RAN \\\cite{RAN}\end{tabular} & \begin{tabular}[c]{@{}c@{}} OADN \\\cite{OADN}\end{tabular}               & \begin{tabular}[c]{@{}c@{}} Wang's \\\cite{Wangs}\end{tabular}           & \begin{tabular}[c]{@{}c@{}} Latent\\-OFER\end{tabular}                 \\ \hline
FED-RO & 66.5           & 69.3           & 70.5           & 68.0         & 68.1 & \color{blue}{70.0}  & {\color{red} 71.8} \\ \hline
\end{tabular}
}
\vspace{3pt}
\caption{Accuracy (\%) comparison to the state-of-the-art results on the FED-RO dataset.}
\vspace{-10pt}
\end{table}

Tables 5 and 6 present the FER performance for occluded images in AffectNet and RAF-DB, where the proposed Latent-OFER model achieved an accuracy of 66.5\% and 84.2\% with the default setting (2.5\%$p$ and 1.5\%$p$ better than OADN \cite{OADN} and RAN \cite{RAN}), respectively.

\begin{table}
\centering
\resizebox{0.9\columnwidth}{!}{%
\begin{tabular}{|c|c|c|c|c|}
\hline
                    & \begin{tabular}[c]{@{}c@{}} BoostGAN\\\cite{Boostgan}\end{tabular} & \begin{tabular}[c]{@{}c@{}} RAN \\\cite{RAN}\end{tabular} & \begin{tabular}[c]{@{}c@{}} OADN \\\cite{OADN}\end{tabular}               & \begin{tabular}[c]{@{}c@{}} Latent\\-OFER\end{tabular}                 \\ \hline
Occlusion-AffectNet & 43.4              & 58.5         & {\color{blue} 64.0} & {\color{red} 66.1} \\ \hline
\end{tabular}
}
\vspace{3pt}
\caption{Accuracy (\%) comparison to the state-of-the-art results on the occlusion-AffectNet dataset.}
\end{table}

\begin{table}
\centering
\resizebox{0.9\columnwidth}{!}{%
\begin{tabular}{|c|c|c|c|c|}
\hline
                 & \begin{tabular}[c]{@{}c@{}}BoostGAN\\ \cite{Boostgan}\end{tabular} & \begin{tabular}[c]{@{}c@{}}RAN\\ \cite{RAN}\end{tabular} & \begin{tabular}[c]{@{}c@{}}Wang's\\ \cite{Wangs}\end{tabular} & \begin{tabular}[c]{@{}c@{}}Latent\\-OFER\end{tabular} \\ \hline
Occlusion-RAF-DB & 55.4              & {\color{blue} 82.7} & 82.5 & {\color{red} 84.2} \\ \hline
\end{tabular}
}
\vspace{3pt}
\caption{Accuracy (\%) comparison to the state-of-the-art results on the occlusion-RAF-DB dataset.}
\end{table}

Table 7 compares the accuracy and complexity of the OFER models. The complexity is estimated by ourselves. As demonstrated in Table 7, OFER models exhibit poor performance on typical FER datasets. This limitation makes them unsuitable for general use unless the unoccluded image network-assisted approach described in Section 1 is employed. In contrast, the proposed model demonstrates acceptable performance on typical and occluded FER datasets. Specifically, the proposed model achieves accuracy rates of 63.9\% and 89.6\%, which represent improvements of 2.0\%$p$ and 2.4\%$p$, respectively, over OADN.
\vspace{-10pt}

\begin{table}
\scriptsize
\centering
\begin{tabular}{|c|cc|cc|}
\hline
                         & \multicolumn{2}{c|}{Accuracy}                             & \multicolumn{2}{c|}{Complexity}               \\ \cline{2-5} 
\multirow{-2}{*}{Method} & \multicolumn{1}{c|}{AffectNet}                   & RAF-DB & \multicolumn{1}{c|}{Flops(G)} & Parameters(M) \\ \hline
gACNN \cite{attention_cnn}                   & \multicolumn{1}{c|}{58.8}                        & 85.1   & \multicolumn{1}{c|}{331}      & 370           \\ \hline
WGAN AE  \cite{WGAN}                & \multicolumn{1}{c|}{59.7}                        & 83.5   & \multicolumn{1}{c|}{353}      & 400           \\ \hline
Pan's   \cite{Pans}                 & \multicolumn{1}{c|}{57.1}                        & 80.2   & \multicolumn{1}{c|}{5}        & 25            \\ \hline
Xia's   \cite{Xias}                   & \multicolumn{1}{c|}{57.5}                        & 82.7   & \multicolumn{1}{c|}{230}      & 360           \\ \hline
RAN   \cite{RAN}                   & \multicolumn{1}{c|}{52.6}                        & 86.9   & \multicolumn{1}{c|}{390}      & 300           \\ \hline
OADN \cite{OADN}                    & \multicolumn{1}{c|}{{\color{blue} 61.9}} & {\color{blue}87.2}   & \multicolumn{1}{c|}{122}      & 250           \\ \hline
Wang's  \cite{Wangs}                 & \multicolumn{1}{c|}{60.2}                        & 86.0   & \multicolumn{1}{c|}{102}      & 210           \\ \hline
Latent-OFER              & \multicolumn{1}{c|}{\color{red} 63.9}   & {\color{red} 89.6}   & \multicolumn{1}{c|}{156}      & 373           \\ \hline
\end{tabular}
\vspace{5pt}
\caption{Accuracy (\%) and complexity comparison to the occluded FER models on the typical FER datasets} %
\vspace{-15pt}
\end{table}

\section{Ablation study}

This section presents the experiments investigating the effects of several modules on FER performance

\textbf{Effect of self-assembly layer.} We use RAF-DB test dataset and synthesize irregular masks \cite{mask} for each image to make comparison. We replaced the self-assembly layer with a conventional 3$\times$3 layer and the CSA layer \cite{liu2019coherent}. As presented in Table 8, we used standard evaluation metrics, the PSNR \cite{psnr_ssim} and SSIM \cite{psnr_ssim}, to quantify the module performance. Moreover, as illustrated in Figure 8, the mask part fails to reconstruct reasonable content when we used conventional 3$\times$3 layer. Although the CSA \cite{liu2019coherent} can improve the performance compared to conventional. Table 9 reveals that the reconstruction results of CSA inpainting still lack facial expression attributes. Table 9 shows the average accuracy of facial expression prediction with images regenerated by MAE, CSA, and hybrid reconstruction network. For fair comparison, the same expression recognition network and the same partially masked test set were used. Compared with them, the proposed method performs better. The effect of semantic consistency loss is included in the Supplementary Material.
\vspace{10pt}

\begin{table}
\centering
\scriptsize
\begin{tabular}{|c|c|c|}
\hline
Method        & PSNR($\uparrow$) & SSIM($\uparrow$) \\ \hline
Conventional  & 26.36                         & 0.868                         \\ \hline
CSA           & 25.48                         & 0.860                         \\ \hline
Self-assembly & 26.65                         & 0.880                         \\ \hline

\end{tabular}%
\vspace{5pt}
\caption{Quantitative comparisons results between conventional, CSA \cite{liu2019coherent}, and self-assembly.}
\vspace{5pt}
\end{table}

 \begin{figure}
 \vspace{-15pt}
\begin{center}
\includegraphics[width=0.95\linewidth]{"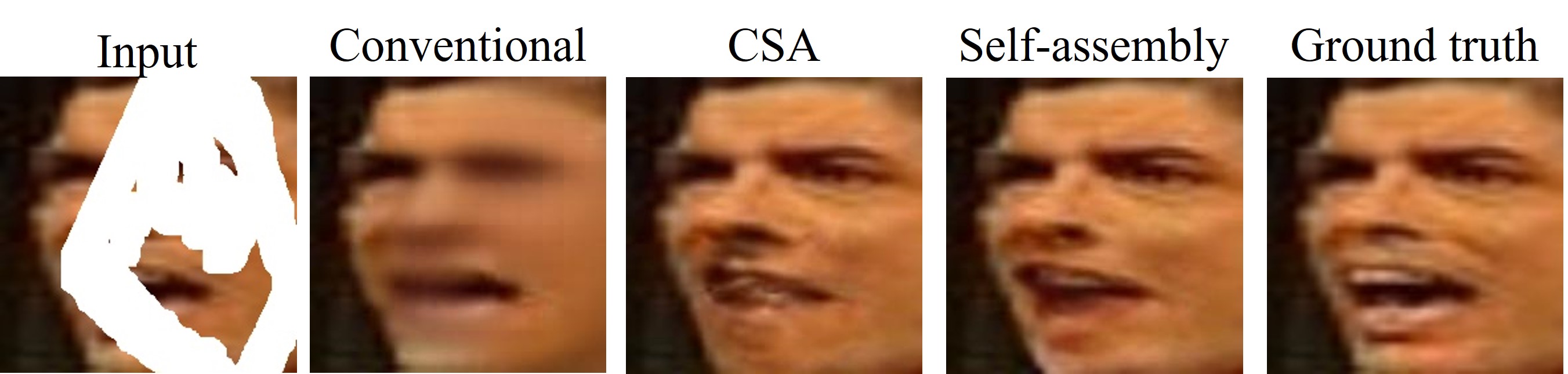"}
\end{center}
\vspace{-8pt}
\caption{Effect of the self-assembly layer. Conventional and CSA \cite{liu2019coherent} are results of the proposed module which replaces the self-assembly layer with the conventional and CSA layers respectively.}
\end{figure}
\vspace{-8pt}
\begin{table}
\centering
\resizebox{0.9\columnwidth}{!}{%

\begin{tabular}{|c|c|c|c|}
\hline
            & MAE \cite{MAE}                         & CSA \cite{liu2019coherent}                          & Hybrid reconstruction\\ \hline
\begin{tabular}[c]{@{}c@{}}Reconstructed-\\RAF-DB \end{tabular} & 72.6 & 71.6 & 77.3 \\ \hline
\end{tabular}}
\vspace{2pt}
\caption{Accuracy (\%) comparison of FER using irregular masked facial image reconstruction results by MAE, CSA, Hybrid reconstruction network.}
\vspace{-15pt}
\end{table}

\textbf{Effect of occlusion detection and reconstruction module.} We demonstrated the effectiveness of the deocclusive auto-encoder with occluded KDEF images created through synthesis. As presented in Table 10, image reconstruction is not performed when the occlusion detector is not used, leading to low FER performance. However, even without the proposed module, the proposed model performed well compared to typical models.
\vspace{4pt}

\textbf{Effect of latent vectors and expression-relevant feature extractor.} As shown Table 10, this study evaluates the efficiency of the deocclusive autoencoder and expression-related feature extractor in FER. The results reveal that using only the ViT-latent vector extracted from occluded data results in poor performance. Although using only CNN features improved the performance, it remained unacceptable. However, training with the ViT-latent vectors and CNN features improves performance by about 4.0\%$p$ compared to only CNN features. Additionally, performance improves by 8.2\%$p$ when images are deoccluded via a hybrid reconstruction network. 

These results indicate that all proposed modules contribute to performance improvement, and the best performance is achieved when CNN features and expression-relevant ViT-latent vectors are trained cooperatively.
\vspace{-5pt}

\begin{table}
\centering

\resizebox{0.9\columnwidth}{!}{%
\begin{tabular}{|c|c|c|c|c|}
\hline
\begin{tabular}[c]{@{}c@{}}Image\\ recon-\\ struction\end{tabular} & \begin{tabular}[c]{@{}c@{}}CNN-\\ features\end{tabular} & \begin{tabular}[c]{@{}c@{}}Full\\ ViT-latent\\ vectors\end{tabular} & \begin{tabular}[c]{@{}c@{}}Extracted\\ ViT-latent\\ vectors\end{tabular} & Accuracy (\%) \\ \hline
\textbf{}      &   & \checkmark   &   & 15.2 \\\hline
&       &         & \checkmark       & 20.1 \\\hline
     & \checkmark    &        &   & 75.4 \\ \hline
  & \checkmark  & \checkmark    &     & 76.5 \\ \hline
  & \checkmark     &      & \checkmark   & 78.5 \\ \hline
\checkmark    &   & \checkmark  &      & 64.9 \\\hline
\checkmark &  &  & \checkmark& 66.4 \\\hline
\checkmark  & \checkmark &  &    & 82.7 \\ \hline 
\checkmark  & \checkmark & \checkmark  &   & 84.2 \\\hline
\checkmark  & \checkmark  & & \checkmark  & 86.7 \\ \hline
\end{tabular}%
}
\vspace{3pt}
\caption{Module evaluation in Latent-OFER on Syn-KDEF.}
\vspace{-15pt}
\end{table}
\vspace{-3pt}
 \section{Discussion}
This section discusses two topics. First, the performance of the occluded patch detector can be considered dominant. The proposed model restores images and extracts latent vectors to use for FER. In cases where occlusion patch detection fails to perform adequately, latent vectors are not extracted correctly because the image is not fully restored. However, the model uses spatial attention to extract expression-relevant features. In images with poor occlusion patch detection, the corresponding patch has lower weights As a result, the influence of poor detection on the overall performance is mitigated. The second is the potential limitations of scalability when using image reconstruction modules trained with a single dataset. These models may not be adaptable enough to handle a variety of datasets, thereby making it difficult to reconstruct a complete image. To overcome this challenge, extra training on diverse datasets or the adoption of uniform alignment across datasets is necessary. Therefore, to enhance the flexibility of image reconstruction modules, a multi-dataset training approach or standardized alignment methods can be implemented.

 \section{Conclusion}
This paper addresses FER in real-world occlusion. Our method involves detecting the occluded parts of the face and then reconstructing them using a specially designed reconstruction network to produce images as if they were unoccluded. In addition, expression-relevant latent vectors were extracted and learned cooperatively with the CNN features. The proposed Latent-OFER model works well under occluded and unoccluded conditions. We evaluate this method on occluded FER datasets and typical FER datasets, and the proposed method achieves state-of-the-art accuracy.

\end{document}